\documentclass{article}

\usepackage[preprint]{neurips_2026}

\usepackage[utf8]{inputenc}
\usepackage[T1]{fontenc}
\usepackage{hyperref}
\usepackage{url}
\usepackage{booktabs}
\usepackage{amsfonts}
\usepackage{nicefrac}
\usepackage{microtype}
\usepackage{xcolor}
\usepackage{graphicx}
\usepackage{amsmath}
\usepackage{amssymb}
\usepackage{multirow}
\usepackage{algorithm}
\usepackage{algorithmic}
\usepackage{subcaption}
\usepackage{enumitem}
\usepackage{wrapfig}

\newcommand{\cmark}{$\checkmark$}
\newcommand{\xmark}{$\times$}

\title{Beyond Scalar Distances: Semantic Attribute Gradients from Frozen MLLMs for Visual Embeddings}

\author{%
  Shubhang Bhatnagar\thanks{Equal contribution.} \quad
  Dheeraj Baiju\footnotemark[1] \quad
  Narendra Ahuja \\
  University of Illinois Urbana-Champaign \\
  \texttt{sb56@illinois.edu} \quad
  \texttt{dheerajbaiju501@gmail.com} \quad
  \texttt{n-ahuja@illinois.edu}
}

\begin{document}

\maketitle

\begin{abstract}

Vision encoders for retrieval are typically trained with class-label supervision: each training pair reduces to a scalar that uniformly pushes the embedding apart or pulls it together, as if every visual attribute either differed or matched. A multimodal large language model (MLLM), shown the same pair, can articulate those attributes and use them to predict whether the images share a class. We propose \textbf{SAGA}, a framework that turns this language-grounded, attribute-aware perception into a training signal for the encoder itself.  Specifically, we use Group Relative Policy Optimization (GRPO) to reward the MLLM for correct predictions on the vision encoder's tokens. Since correct predictions require those tokens to expose the specific attributes that differ or match between the pair, the gradient pushes the encoder to encode them, replacing the uniform pair-level scalar with attribute-resolved supervision. An auxiliary attention-distillation loss anchors the encoder's embedding to tokens the MLLM attended to, and a standard metric-learning loss shapes the embedding geometry for nearest-neighbour retrieval. The MLLM is frozen throughout and discarded at inference, matching the deployment cost of a metric-learning baseline. SAGA improves Recall@1 by 3 to 6 points over state-of-the-art baselines on CUB-200-2011, Cars-196, FGVC-Aircraft, and iNaturalist Aves on zero-shot image retrieval.

\end{abstract}

\section{Introduction}
\label{sec:intro}

\begin{figure}
    \centering
    \includegraphics[width=\linewidth]{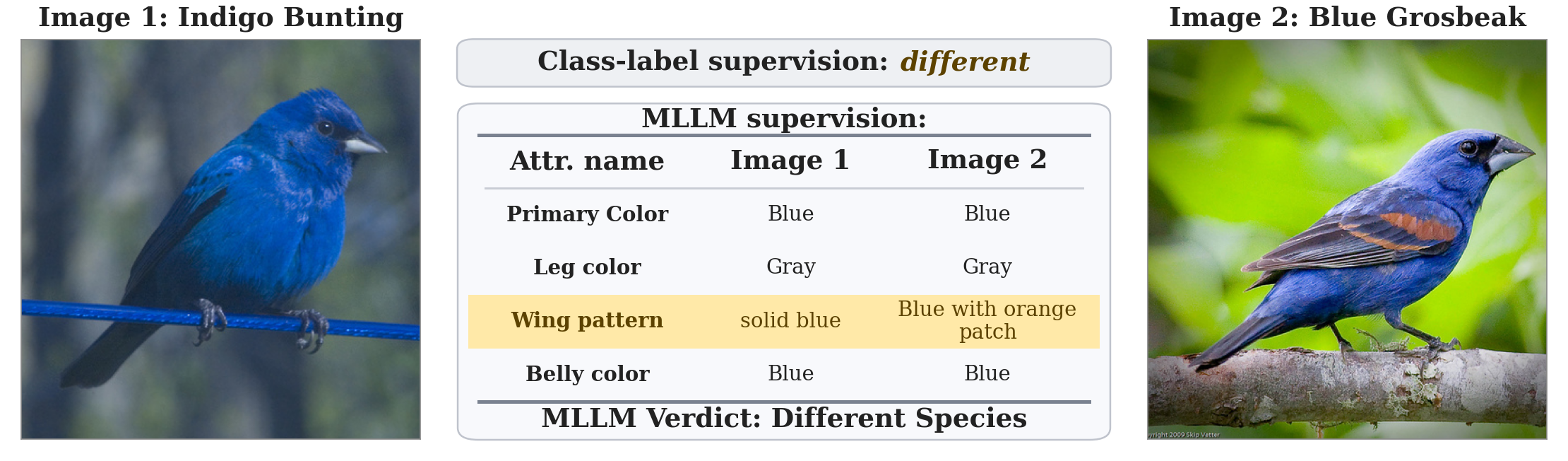}
    \caption{\textbf{Using only class labels for  images reduces supervision to a scalar, whereas an MLLM resolves it into attributes.}
A class-label loss collapses the difference  between two very similar-looking bird species into a single `different' scalar, pushing every embedding dimension apart, even those potentially encoding shared attributes like blue plumage and leg color. A frozen MLLM, by contrast, can identify which attributes match and include them in reaching the same-/different-species verdict. Our method, SAGA, harnesses this by rewarding correct verdicts and reinforcing precisely those feature components (directions) that the MLLM's discrimination relies on, while leaving shared-attribute directions untouched.}
\vspace{-10pt}
\label{fig:teaser} 
\end{figure}

A visual encoder must embed images along the dimensions that distinguish them: the shape of a bill, the pattern of a wing, the silhouette of a garment, the geometry of a tail. The dominant paradigm (metric learning) trains them with class labels alone~\citep{chopra2005learning,multisimilarity_dml,proxy_nca,bhatnagar2025potentialfield}, a binary signal that acts on every attribute in unison, pulling all of them together when classes match and pushing all of them apart when classes differ, even when two images share most attributes and are distinguished by only a few. This is the wrong inductive bias for zero-shot image retrieval~\citep{sopdataset,proxy_anchor}, where test classes come from a disjoint label set and are separated by attribute combinations the training signal never required the encoder to represent. Figure~\ref{fig:teaser} makes this concrete: an Indigo Bunting and a Blue Grosbeak (from the CUB200 \cite{cubdataset} dataset)  share a deep-blue plumage and gray legs, differing only in their wing bars. A class-label scalar reduces this pair to a uniform ``different,'' carrying no information that those few attributes are the ones that matter while the rest agree.

Multimodal large language models (MLLMs) trained on image-text data~\citep{liu2024visual,bai2025qwen3vl,chen2024internvl,grattafiori2024llama} acquire exactly this perception of visual structure. Asked about an image, an MLLM articulates fine-grained attributes (shapes, patterns, textures, structural proportions) and localizes them on the image while it reasons. For the pair in Figure~\ref{fig:teaser}, we can see that the MLLM identifies \emph{wing pattern: two orange wing bars} for Bird~2 against \emph{wing pattern: solid blue} for Bird~1, and concludes 'different species'. We ask whether the MLLM's sensitivity to detail can serve as a training-time supervisor for a visual encoder, turning its high emphasis on certain attributes into learning gradients that reshape the encoder's embedding space.

In this work, we answer this affirmatively and propose \textbf{SAGA} (\textbf{S}emantic \textbf{A}ttribute \textbf{G}radients from \textbf{A}djudication), a framework that turns a frozen MLLM into a training-time supervisor for the visual encoder of a retrieval system. Our visual encoder is the vision tower of a multimodal LLM, which emits a sequence of patch tokens fed to an MLLM which is asked to compare image pairs by describing their attributes ~\citep{wei2022cot}. Correct same/different class verdicts are rewarded via Group Relative Policy Optimization (GRPO)~\citep{shao2024deepseekmath}; the resulting gradient flows back through the frozen language backbone into the encoder, pushing it to represent the discriminative attributes the MLLM had to perceive correctly to reach those verdicts. In Figure~\ref{fig:teaser}, the correct `different species' verdict of the MLLM relies on it identifying Bird 2's orange wing bars, so the policy gradient reinforces the encoder to identify and discriminate this attribute, while directions encoding the shared blue plumage and gray legs receive no such reinforcement.

Encoded discriminative attributes still have to be aggregated into a single retrieval vector, so we attach a small pooler that collapses the patch tokens into the embedding used for nearest-neighbor search at inference. The same forward pass also reveals which image regions/tokens the MLLM attended to while reasoning about each image's attributes. We distill~\citep{hinton2015distilling,zagoruyko2017paying} this attention into the pooler, a lightweight module that aggregates the encoder's output tokens for an image into a single embedding vector used for nearest-neighbor retrieval at inference. Without this signal the pooler would be left to discover attribute-relevant tokens from class labels alone, the same statistical-discovery problem identified above for the encoder.


The MLLM is frozen throughout training and  is used only at training time to produce GRPO rewards and attention targets. Once training is complete only the vision encoder and pooler are retained for deployment. Retrieval is performed by only these two components, matching the deployment cost of any standard metric learning pipeline. The supervisory signal requires only the pairwise class labels already used by the metric learning objective; no attribute annotations are needed.


Our main contributions are:
\begin{itemize}
\item We present SAGA, a framework that uses a frozen multimodal LLM as a training-time supervisor to learn a visual encoder using attribute-aware supervision gradients that go beyond the scalar pairwise similarity between class labels.
\item We train the encoder with a reinforcement learning objective that rewards the encoder when the MLLM correctly judges image pairs as being from the same class or different classes, and distill the MLLM's attention into the pooler. This (1) teaches the encoder to incorporate in its representation the attributes the MLLM uses to judge, and (2) teaches the pooler to weight the image regions the MLLM attends to when forming its verdict.


\item We evaluate SAGA on four zero-shot image retrieval benchmarks (CUB-200-2011, Cars-196, FGVC-Aircraft, iNaturalist Aves), where it improves Recall@1 by 3--6\% over state-of-the-art baselines on the same vision backbone.
\end{itemize}

\section{Related Work}
\label{sec:related}
\paragraph{Deep metric learning.}
Deep metric learning (DML) is the dominant framework for training a vision encoder whose output space is itself a semantic geometry: standard distance metrics over embeddings recover class-level similarity, and the resulting features are intended to generalize to disjoint test classes for downstream tasks such as zero-shot image retrieval~\citep{sopdataset,proxy_anchor} and face verification~\citep{schroff_facenet_2015}. Methods in this family supervise the encoder with a scalar pairwise objective derived from class labels, instantiated either through tuple-based losses operating directly on samples (contrastive~\citep{chopra2005learning,hadsell2006dimensionality}, triplet~\citep{schroff_facenet_2015}, multi-similarity~\citep{multisimilarity_dml}) or through proxy-based losses that replace tuple mining with learnable class representatives (Proxy-NCA~\citep{proxy_nca}, Proxy-Anchor~\citep{proxy_anchor}, HIER\cite{kim2023hier}, DDML \cite{park2025deep} Potential Field~\citep{bhatnagar2025potentialfield}). 
Across both families, supervision per pair reduces to a single scalar that acts on every attribute dimension in unison, telling the encoder that two images should move closer or farther but not which visual attributes carry the class signal.

\paragraph{Multimodal large language models.}
MLLMs trained on image-text data, e.g., LLaVA~\citep{liu2024visual}, Qwen-VL~\citep{bai2025qwen3vl}, and InternVL~\citep{chen2024internvl}, articulate fine-grained visual attributes through language and localize them on the image while reasoning. Group Relative Policy Optimization (GRPO)~\citep{shao2024deepseekmath} has become the standard recipe for aligning these models with non-differentiable rewards, including grounded visual reasoning~\citep{fan2025gritteachingmllmsthink,wang2026traceableevidenceenhancedvisual} and reconstructive encoder objectives~\citep{yan2026unifiedmultimodalmodelsautoencoders}. These works fine-tune the MLLM for VQA or grounded reasoning as a whole. SAGA uses GRPO in the opposite role, treating the frozen MLLM as a loss function whose policy gradient supervises a retrieval encoder.
A complementary line observes that an MLLM's internal attention often localizes salient regions even when its textual output is flawed~\citep{hou2025visionlanguagemodelsreallyunderstand}, and exploits this at \emph{inference} time by reallocating resolution or tokens toward attended regions~\citep{dalal2025constructive,zhang2025mllms}; SAGA distills that same attention at \emph{training} time into a retrieval pooler.

\paragraph{Language-guided visual representation learning.}
A complementary thread uses textual descriptions as supervision for visual representations. CLIP~\citep{radford2021learning} and SigLIP~\citep{zhai2023sigmoid} align image and text embeddings via contrastive pretraining, and subsequent work adapts these models with LLM-generated class descriptions for zero-shot recognition~\citep{menon2022visualclassificationdescriptionlarge,saha2024improved}; CAP-FGVC~\citep{schmidt2025saccadicvisionfinegrainedvisual} extends the idea to fine-grained retrieval with caption-supervised contrastive losses. Similar to DML methods, these consume language as fixed targets that align embeddings to text while also requiring caption level labels for such fine-grained images. SAGA does not need such caption level supervision, and only using the class label supervision and GRPO can make encoder learn features about discriminative attributes

\section{Method}
\label{sec:method}

\subsection{Setup and Notation}
\label{sec:method_setup}
Deep metric learning (DML) learns a semantic distance over images from a labelled dataset $\mathcal{D} = \{(\textbf{I}_i, y_i)\}_{i=1}^{|\mathcal{D}|}$ with $y_i \in \{1,\dots,N\}$, parameterizing an image-to-embedding map $g_{\theta,\phi}: \textbf{I} \mapsto \textbf{z} \in \mathbb{R}^{D_e}$ and taking $d(\textbf{I}_1, \textbf{I}_2) = \|\textbf{z}_1 - \textbf{z}_2\|_2$; $d$ should be small for same-class pairs and large otherwise.


\paragraph{Vision encoder and retrieval pooler.}
We factor $g_{\theta,\phi} = c_\phi \circ f_\theta$ into a vision encoder (vision tower of Qwen3-VL~\citep{bai2025qwen3vl}) and a retrieval pooler $c_\phi$, producing a sequence of patch tokens $\textbf{X} = f_\theta(\textbf{I}) \in \mathbb{R}^{N_p \times D}$ for an image $\textbf{I}$. 
The pooler $c_\phi$ aggregates these into a compact embedding $\textbf{z} = c_\phi(\textbf{X}) \in \mathbb{R}^{D_e}$, instantiated as mean, max, or attention-pooling; ... we write $\boldsymbol{\beta} \in \Delta^{N_p}$ (the probability simplex over the $N_p$ patches) for the pooler's spatial weights when attention-pooling. Both $\theta$ and $\phi$ are trainable.

\paragraph{Frozen MLLM-guided supervision}
To enrich $f_\theta$ with semantic reasoning, we use it as the visual front-end for the language backbone $p_\psi$ of the MLLM. Given $\textbf{X}$ and a text prompt, $p_\psi$ autoregressively generates output tokens; we use $\boldsymbol{\alpha}^{(t)} \in \Delta^{N_p}$ for its attention distribution at an intermediate decoder layer (justified empirically in Sec.~\ref{sec:kl_loss}) from a generated token $t$ over the patch positions of $\textbf{X}$. The parameters $\psi$ are frozen, but gradients flow through $p_\psi$ into $\theta$. The MLLM is used only at training time and discarded at inference, so the embedding model has the same cost as a standard DML pipeline. For proxy-based DML losses, we additionally maintain $M$ trainable proxies per class, $\textbf{p}_{j,k} \in \mathbb{R}^{D_e}$ for $j \in \{1,\dots,N\}, k \in \{1,\dots,M\}$.



\begin{figure}[t]
    \centering
    \includegraphics[width= \textwidth]{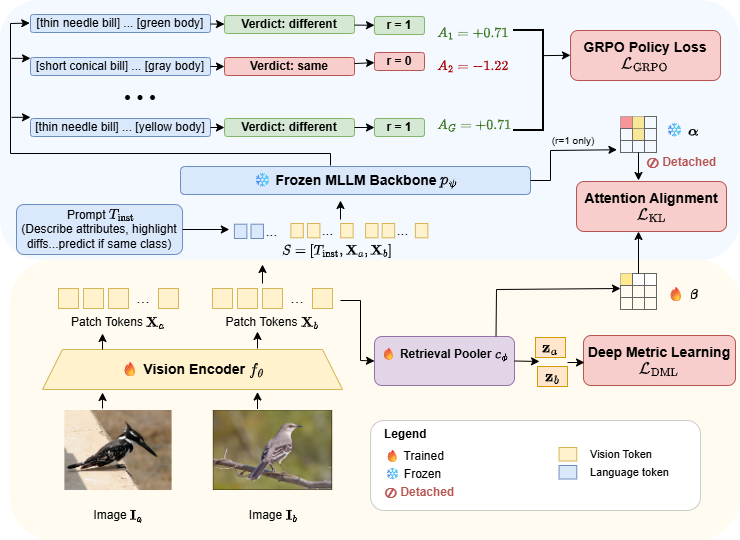}
    \caption{\textbf{Overview. }SAGA uses a frozen MLLM as an attribute-aware supervisor for deep metric learning.
For an image pair $(\mathbf{I}_a, \mathbf{I}_b)$, the trainable vision encoder $f_\theta$ produces patch tokens $\mathbf{X}_a, \mathbf{X}_b$, which feed three losses with complementary roles.
(1)~The tokens and a comparison prompt $T_\text{inst}$ are fed to the frozen MLLM $p_\psi$, which samples $G$ responses ending in a same/different-class verdict; the GRPO loss $\mathcal{L}_\text{GRPO}$ rewards correct verdicts and back-propagates through $p_\psi$ into $f_\theta$, pushing it to encode the \emph{discriminative} attributes the MLLM relied on when making correct predictions. (2)~The MLLM's attention $\alpha$ from the same forward pass reveals which patch tokens the MLLM attended to while describing each image's attributes; on correct rollouts, the per-image mean attribute-attention $\bar{\alpha}$ is distilled into the pooler's attention $\beta$ via $\mathcal{L}_\text{KL}$, encouraging $c_\phi$ to pool over those regions when forming embeddings $z_a, z_b$.  (3)~The pooler $c_\phi$ aggregates the tokens into embeddings $\mathbf{z}_a, \mathbf{z}_b$, which a deep metric learning loss $\mathcal{L}_\text{DML}$ shapes for nearest-neighbor search. The MLLM is frozen throughout and discarded at inference.}
\label{fig:overview}
\vspace{-10pt}
\end{figure}

\subsection{GRPO Attribute Reasoning Loss}
\label{sec:grpo}
The core contribution is using a frozen MLLM as a differentiable, attribute-aware loss function via Group Relative Policy Optimization (GRPO)~\cite{shao2024deepseekmath}. Rather than supervised fine-tuning with fixed targets, the MLLM generates freely, and we reward only the final same/different-class verdict. This reward is computed against the ground-truth class labels $y$ used by the DML loss, adding no extra annotation burden. Intermediate attribute descriptions serve as an implicit chain-of-thought, making the resulting gradient attributionally rich despite the binary reward.

\paragraph{Input construction.}
Given two images $\textbf{I}_a, \textbf{I}_b$ from $\mathcal{D}$ with labels $y_a, y_b$, we construct an input sequence by concatenating their patch tokens with a structured text prompt $T_\text{inst}$:

\begin{equation}
    S = [\, \textbf{X}_a,\;\; \textbf{X}_b,\;\; T_\text{inst} \,].
    \label{eq:input_seq}
\end{equation}
$T_\text{inst}$ instructs the MLLM to: (1)~describe visual attributes in JSON, (2)~highlight key differences, and (3)~predict if the images share a class.

\paragraph{Rollout and reward.}
For each pair, we sample $G$ completions without gradients:
\begin{equation}
    \hat{Y}^{(g)} = (y_1^{(g)}, \dots, y_{T_g}^{(g)}) \sim p_\psi(\cdot \mid S), \quad g = 1, \ldots, G,
\end{equation}
where $T_g$ is the length of the $g$-th completion.
We parse each completion for the verdict field and assign a binary reward:
\vspace{-2pt}
\begin{equation}
    r^{(g)} = \begin{cases} 1 & \text{if the parsed verdict matches } \mathbb{1}[y_a = y_b], \\ 0 & \text{otherwise (including unparseable outputs).} \end{cases}
    \label{eq:reward}
\end{equation}

\paragraph{Policy gradient update.}
We compute group-normalized advantages $A^{(g)} = (r^{(g)} - \bar{r}) / (\sigma_r + \epsilon)$ for each completion. Pairs with $\sigma_r = 0$ contribute no policy-gradient signal and are skipped, with additional pairs drawn from subsequent micro-batches to maintain a target contributing-pair count per optimizer step (DAPO Dynamic Sampling~\citep{yu2026dapo}).
Because each rollout is generated and consumed within a single gradient update, the GRPO importance ratio $\rho_t = \pi_\theta(y_t)/\pi_{\theta_{\text{old}}}(y_t) \equiv 1$ at update time, the surrogate's clip is vacuous, and we use no reference policy ($\beta = 0$), so the GRPO loss reduces to its first-order, advantage-weighted negative log-likelihood form:
\begin{equation}
    \log \pi_{\theta}(y_t^{(g)} \mid y_{<t}^{(g)}, S) \quad \text{for each generated token } t.
\end{equation}
The GRPO loss is the advantage-weighted negative log-likelihood over all generated tokens:
\begin{equation}
    \mathcal{L}_\text{GRPO} = -\frac{1}{|\mathcal{T}|} \sum_{g} \sum_{t=1}^{T_g} A^{(g)} \log \pi_{\theta}(y_t^{(g)} \mid y_{<t}^{(g)}, S),
    \label{eq:grpo_loss}
\end{equation}
where $|\mathcal{T}| = \sum_g T_g$ is the total number of generated tokens across contributing completions. 
Token-level normalization~\citep{yu2026dapo} prevents short completions from dominating the gradient.

\paragraph{Why GRPO provides attribute-aware gradients.}
The policy gradient at each token $\pi(y_t)$ sends signals back to $\theta$ through $\partial \log \pi(y_t) / \partial \mathbf{X} \cdot \partial \mathbf{X} / \partial \theta$, exciting only the dimensions of $\mathbf{X}$ used to predict $y_t$.  Shared-attribute tokens are produced with similar probabilities across correct
and incorrect rollouts: the visual signal for these attributes is identical in $\mathbf{I}_a$ and $\mathbf{I}_b$ by definition, so the MLLM's belief about them is fixed by perception and does not track the verdict outcome. $\log\pi(y_t)$ is therefore roughly constant in $g$, and the advantage-weighted sum vanishes by the mean-zero property of group-normalised advantages.
Discriminating-attribute tokens, in contrast, force the MLLM to commit
to one description per rollout (e.g., ``orange wing bars'' vs.\ ``solid blue'') on the basis of whatever signal $\mathbf{X}$ exposes; if the encoder has not yet cleanly encoded that signal, sampled tokens vary across rollouts and align with the verdict outcome, with rollouts that picked the correct attribute receiving $r^{(g)}=1$ and the others $r^{(g)}=0$. The advantage-weighted sum is therefore non-zero on exactly
these tokens, flowing into the $\mathbf{X}$-directions that resolve them. As those directions sharpen, the MLLM grows confident, rollout disagreement shrinks, and the gradient decays, producing an automatic curriculum onto attributes the encoder has not yet learned. This mirrors the outcome-only credit-assignment mechanism by which DeepSeek-R1~\citep{guo2025deepseek} elicits emergent reasoning from binary correctness rewards.

\subsection{Deep Metric Learning Loss}
\label{sec:dml_loss}
$\mathcal{L}_\text{GRPO}$ shapes which visual signal $f_\theta$ encodes, but does not arrange the resulting embeddings $\textbf{z} = c_\phi(f_\theta(\textbf{I})) \in \mathbb{R}^{D_e}$ for nearest-neighbor search. We therefore retain a standard deep metric learning loss $\mathcal{L}_\text{DML}$ computed over the pooled embeddings of the full training batch, which back-propagates into both $\theta$ and $\phi$ to provide geometric supervision every step.

The two losses are deliberately complementary. $\mathcal{L}_\text{DML}$ decides \emph{where} points sit in $\mathcal{Z}$, while $\mathcal{L}_\text{GRPO}$ decides \emph{which visual signal} the encoder uses to place them there. Removing $\mathcal{L}_\text{DML}$ would leave the GRPO gradient un-anchored to any explicit metric structure; removing $\mathcal{L}_\text{GRPO}$ would leave the geometric supervision attribute-blind, recovering the coarse pairwise signal that motivated this work. Our framework is agnostic to the specific DML objective, and we evaluate three representative variants (InfoNCE~\citep{oord2018representation}, Proxy-Anchor~\citep{proxy_anchor}, and Potential Field~\citep{bhatnagar2025potentialfield}) to demonstrate that the GRPO supervision composes with both proxy-free and proxy-based metric learning.

\subsection{Attention Alignment Loss}
\label{sec:kl_loss}

$\mathcal{L}_{\text{GRPO}}$ updates $f_\theta$ via the frozen LLM, but the pooler $c_\phi$ only perceives these gradients indirectly. To prevent $c_\phi$ from weighting non-discriminative regions and erasing attribute information made encodable in $f_\theta$, we introduce an attention-alignment loss. This supervises the pooler's spatial focus by distilling, on correct rollouts only, the MLLM's internal attention during attribute generation. 

While a final-layer teacher is intuitive, Qwen3-VL-8B’s last-layer attention is dominated by "attention-sink" and register-token artifacts, yielding maps poorly aligned with visual attributes. Using the AttWarp~\cite{dalal2025constructive} framework, we find layer $\ell = 26$ provides the best trade-off, consistently highlighting attribute-relevant regions

Concretely, let $\mathcal{A}_a, \mathcal{A}_b$ denote the attribute-description tokens describing $\mathbf{I}_a$ and $\mathbf{I}_b$, and $\boldsymbol{\alpha}_a^{(t)}, \boldsymbol{\alpha}_b^{(t)} \in \Delta^{N_p}$ denote the head-averaged attention at layer $\ell = 26$ from token $t$, renormalized over patches. We aggregate these into a \emph{mean attribute-attention map} per image:
\begin{equation}
\bar{\boldsymbol{\alpha}}_a \;=\; \frac{1}{|\mathcal{A}_a|}\sum_{t \in \mathcal{A}_a} \boldsymbol{\alpha}_a^{(t)},
\qquad
\bar{\boldsymbol{\alpha}}_b \;=\; \frac{1}{|\mathcal{A}_b|}\sum_{t \in \mathcal{A}_b} \boldsymbol{\alpha}_b^{(t)},
\label{eq:alpha_mean}
\end{equation}
which represents the union of patch regions the LLM attended to while describing attributes. For each pair with reward $r^{(g)} = 1$, we align these with the pooler’s attentions $\boldsymbol{\beta}_a, \boldsymbol{\beta}_b$ via:
\begin{equation}
\mathcal{L}_{\text{KL}} \;=\; D_{\text{KL}}\!\left(\bar{\boldsymbol{\alpha}}_a \,\|\, \boldsymbol{\beta}_a\right) \;+\; D_{\text{KL}}\!\left(\bar{\boldsymbol{\alpha}}_b \,\|\, \boldsymbol{\beta}_b\right).
\label{eq:kl_loss}
\end{equation}
Eq.~\ref{eq:kl_loss} is gradient-equivalent (in $\boldsymbol{\beta}$) to the per-token average $\frac{1}{|\mathcal{A}_a|}\sum_{t \in \mathcal{A}_a} D_{\text{KL}}(\boldsymbol{\alpha}_a^{(t)} \| \boldsymbol{\beta}_a) + \frac{1}{|\mathcal{A}_b|}\sum_{t \in \mathcal{A}_b} D_{\text{KL}}(\boldsymbol{\alpha}_b^{(t)} \| \boldsymbol{\beta}_b)$, differing only by a $\boldsymbol{\beta}$-independent entropy offset. This formulation is cheaper to compute and targets the parts of the rollout that localize on specific visual regions. Gradients flow only into $\phi$; tokens $\mathbf{X}$ and teacher maps $\boldsymbol{\alpha}$ are detached. This teaches $c_\phi$ \emph{where to look}, complementing $\mathcal{L}_{\text{GRPO}}$’s role in determining \emph{what to encode}.

\subsection{Total Objective and Training}
\label{sec:training}

The overall loss is:
\begin{equation}
    \mathcal{L}_\text{Total} = \lambda_\text{dml} \, \mathcal{L}_\text{DML} + \lambda_\text{lm} \, \mathcal{L}_\text{GRPO} + \lambda_\text{kl} \, \mathcal{L}_\text{KL},
    \label{eq:total_loss}
\end{equation}
where $\lambda_\text{dml}$, $\lambda_\text{lm}$, and $\lambda_\text{kl}$ are scalar loss weights. The three losses play complementary roles: $\mathcal{L}_\text{DML}$ optimizes the embedding geometry via $\theta$, $\phi$, and the proxies $\textbf{p}_{j,k}$ (when present); $\mathcal{L}_\text{GRPO}$ provides attribute-aware gradients to $\theta$ via the frozen LLM; and $\mathcal{L}_\text{KL}$ teaches $\phi$ where to attend.

\paragraph{Per-step training flow.}
Each training step proceeds in three phases:
\textbf{(A)}~compute embeddings for the full batch and apply $\mathcal{L}_\text{DML}$;
\textbf{(B)}~sample image pairs from the batch, run $G$ rollouts per pair through the frozen MLLM, and score binary rewards;
\textbf{(C)}~for pairs with non-zero advantage variance, run the differentiable forward pass and apply $\mathcal{L}_\text{GRPO}$, additionally applying $\mathcal{L}_\text{KL}$ for the rollouts within those pairs that received reward~1.
We use gradient accumulation across pairs within a step, followed by gradient clipping and a single optimizer update. Algorithm~\ref{alg:training} (Appendix~\ref{app:impl_details}) provides full pseudocode.


\subsection{Inference}
\label{sec:inference}

At inference, the frozen MLLM $p_\psi$ is discarded entirely. For a query image $\textbf{I}_q$ and gallery $\mathcal{G} = \{\textbf{I}_g\}$, retrieval is standard nearest-neighbor search in the embedding space:
\begin{equation}
    \textbf{z}_q = c_\phi(f_\theta(\textbf{I}_q)), \quad \textbf{z}_g = c_\phi(f_\theta(\textbf{I}_g)), \quad \text{rank by } \|\textbf{z}_q - \textbf{z}_g\|_2.
\end{equation}
The deployed system consists only of the trained vision encoder and pooler, so its inference cost is identical to any standard DML pipeline; the MLLM serves solely as a training-time supervisor.

\section{Experiments}
\label{sec:experiments}

\subsection{Setup}
\label{sec:setup}

\textbf{Datasets:} We empirically compare our method (SAGA) against state-of-the-art DML baselines on four zero-shot image retrieval benchmarks: \textbf{(1)} the CUB-200-2011 dataset~\citep{cubdataset} consisting of $11{,}788$ images from $200$ bird species, \textbf{(2)} the Cars-196 dataset~\citep{carsdataset} containing $~16$k images from $196$ car model categories, \textbf{(3)} the FGVC-Aircraft dataset~\citep{maji2013aircraft} with $10{,}000$ images from $100$ aircraft variants, and \textbf{(4)} the \textbf{iNat-Aves} benchmark we curate from the iNaturalist-2021 dataset~\citep{vanhorn2021inat}: starting from the \texttt{train\_mini} split (50 img/ species) we retain only the taxonomic class \emph{Aves}, yielding $\sim$$1{,}486$ species and $\sim$$74$k images. 

Classes in all four benchmarks are distinguished by visual attributes that an MLLM can reason about.  We exclude the product-retrieval benchmarks SOP~\citep{sopdataset} and In-Shop~\citep{liu2016deepfashion}, since their classes separate on object identity rather than fine-grained attributes. Per-dataset prompt templates, preprocessing details, and full dataset statistics are reported in Appendix~\ref{app:datasets}.

\textbf{Evaluation Settings:} Following the standard zero-shot retrieval protocol of prior DML work~\citep{sopdataset, proxy_anchor, multisimilarity_dml, bhatnagar2025potentialfield}, classes are partitioned into disjoint train and test halves and the model is evaluated on \emph{unseen} classes at $224 \times 224$ resolution. CUB-200-2011 and Cars-196 use the canonical splits; for FGVC-Aircraft and iNat-Aves we apply the same class-disjoint half-split convention  (first half train, second half test; full details in Appendix~\ref{app:datasets}). We report Recall@$K$ (fraction of queries with a same-class neighbour among the $K$ nearest) and Normalized Mutual Information (NMI), computed between $k$-means cluster assignments on the test embeddings (with $k$ equal to the number of test classes) and the ground-truth labels, capturing both nearest-neighbour and global embedding-space structure.

\textbf{Backbone:} We use \textbf{Qwen3-VL-8B}~\citep{bai2025qwen3vl} as our MLLM for our main results, with its vision tower instantiating the encoder $f_\theta$ and its language backbone serving as the frozen supervisor $p_\psi$. All DML baselines use the same Qwen3-VL-8B vision tower for fair comparison; baselines use mean pooling over patch tokens, while our full method uses our learned attention pooler. The pooler outputs $\ell_2$-normalized embeddings of dimension $D_e = 4096$.

\textbf{Training parameters:} The encoder and pooler are trained with AdamW with cosine-annealed learning rates, GRPO group size $G = 8$, and $P = 8$ balanced same/different-class pairs per step. All experiments use a single NVIDIA H200 (141\,GB) GPU with bfloat16 mixed precision. Full hyperparameter values, sweeps, and ablations of these choices are reported in Appendix~\ref{app:impl_details}.

\subsection{Image Retrieval Performance}
\label{sec:main_results}

\begin{table*}[t]
\centering
\caption{\textbf{Main results on zero-shot image retrieval.} Recall@$1$, Recall@$4$ (\%) and Normalized Mutual Information (NMI, $\in[0,1]$) on four fine-grained benchmarks (CUB-200-2011, Cars-196, FGVC-Aircraft, and our iNat-Aves subset of iNaturalist-2021). All methods share the same Qwen3-VL-8B vision tower; baselines use mean pooling. Baselines: \textbf{PA} = Proxy Anchor~\citep{proxy_anchor}, \textbf{PF} = Potential Field~\citep{bhatnagar2025potentialfield}. Best per column in \textbf{bold}, second-best \underline{underlined}. $\pm$ values for SAGA are standard deviation over 3 random seeds}
\label{tab:main_results}
\small
\setlength{\tabcolsep}{4pt}
\renewcommand{\arraystretch}{1.1}
\resizebox{\textwidth}{!}{%
\begin{tabular}{l *{4}{ccc}}
\toprule
& \multicolumn{3}{c}{\textbf{CUB-200}} & \multicolumn{3}{c}{\textbf{Cars-196}} & \multicolumn{3}{c}{\textbf{Aircraft}} & \multicolumn{3}{c}{\textbf{iNat-Aves}} \\
\cmidrule(lr){2-4} \cmidrule(lr){5-7} \cmidrule(lr){8-10} \cmidrule(lr){11-13}
\textbf{Method} & R@$1$ & R@$4$ & NMI & R@$1$ & R@$4$ & NMI & R@$1$ & R@$4$ & NMI & R@$1$ & R@$4$ & NMI \\
\midrule
Pre-trained backbone               & $75.6$ & $91.8$ & $0.77$ & $70.7$ & $88.5$ & $0.49$ & $53.1$ & $76.1$ & $0.43$ & $42.2$ & $64.8$ & $0.65$ \\
\cmidrule(lr){1-13}
PA                       & $79.5$ & $92.0$ & $0.79$ & $93.4$ & $97.3$ & $0.84$ & $73.1$ & $92.8$ & $0.68$ & $54.1$ & $73.1$ & $0.72$ \\
PF                       & $\underline{81.6}$ & $\underline{92.9}$ & $\underline{0.81}$ & $\underline{93.7}$ & $\underline{97.8}$ & $\underline{0.86}$ & $\underline{77.4}$ & $\underline{93.2}$ & $\underline{0.72}$ & $\underline{55.6}$ & $\underline{75.0}$ & $\underline{0.73}$ \\
\midrule
\textbf{SAGA (ours)}  & $\mathbf{87.9}_{\pm 0.3}$ & $\mathbf{96.3}_{\pm 0.2}$ & $\mathbf{0.83}$ & $\mathbf{97.0}_{\pm 0.3}$ & $\mathbf{98.6}_{\pm 0.1}$ & $\mathbf{0.89}$ & $\mathbf{83.5}_{\pm 0.4}$ & $\mathbf{93.9}_{\pm 0.3}$ & $\mathbf{0.77}$ & $\mathbf{60.1 \pm 0.4}$ & $\mathbf{77.1 \pm 0.3}$ & $\mathbf{0.80}$ \\
\bottomrule
\end{tabular}%
}
\vspace{-5pt}
\end{table*}

As seen in Table~\ref{tab:main_results}, our method significantly outperforms standard DML baselines on all four fine-grained datasets. It outperforms the best-performing baseline, PotentialField~\citep{bhatnagar2025potentialfield}, in terms of Recall@1 (R@1) by $\mathbf{6.3\%}$ on CUB-200-2011, $\mathbf{3.3\%}$ on Cars-196, $\mathbf{6.1\%}$ on FGVC-Aircraft, and $\mathbf{4.5\%}$ on iNat-Aves, and shows similar margins over ProxyAnchor~\citep{proxy_anchor}. The performance gains are largest on the most attribute-driven benchmarks (birds and aircraft variants), consistent with our hypothesis: these classes are distinguished by subtle attributes (bill shape, wing-bars, eye rings for birds; tail and wing geometry for aircraft) that the MLLM explicitly reasons about during GRPO training, rather than by coarse object identity. The substantial gain on iNat-Aves further demonstrates that this attribute-aware supervision scales to a much larger label space ($\sim$$743$ training species). The improvement persists at $R@4$ and in NMI, indicating that the benefit extends to the overall structure of the embedding space. Qualitative comparisons are in Sec.~\ref{sec:qualitative}.


\subsection{Ablation Studies}
\label{sec:ablations}
Unless otherwise stated, all ablations follow the experimental setting of Sec.~\ref{sec:setup}: we use the Qwen3-VL-8B vision tower with our learned attention pooler, train and evaluate on CUB-200-2011 (the most attribute-driven of our four benchmarks) and FGVC-Aircraft (where our main results show the largest absolute R@1 gain), and use the hyperparameters reported in Sec.~\ref{sec:setup} (full values in Appendix~\ref{app:impl_details}). We report Recall@1 on the held-out test classes of both datasets.

\begin{table}[t]
\centering
\begin{minipage}[t]{0.48\textwidth}
\centering
\caption{\textbf{Loss component ablation} on CUB-200-2011 and FGVC-Aircraft (R@1, \%). All configurations include the DML term ($\mathcal{L}_\text{DML}$), instantiated as PF; ticks indicate which losses are added. The indented italic row swaps the per-dataset attribute list for a generic prompt (App.~\ref{app:prompt_generic}) to test prompt sensitivity. Baselines: \textbf{PF} = Potential Field~\citep{bhatnagar2025potentialfield}.}
\label{tab:ablation_components}
\small
\setlength{\tabcolsep}{4pt}
\vspace{2pt}
\begin{tabular}{ccccc}
\toprule
$\mathcal{L}_\text{GRPO}$ & $\mathcal{L}_\text{KL}$ & & CUB & FGVC-A \\
\midrule
\xmark & \xmark & PF only        & 81.6 & 77.4 \\
\cmark & \xmark & + GRPO         & 87.0 & 82.3 \\
       &        & \quad\textit{(generic prompt)} & \textit{84.1} & \textit{79.6} \\
\xmark & \cmark & + KL           & 82.1 & 78.1 \\
\cmark & \cmark & \textbf{SAGA}  & \textbf{87.9} & \textbf{83.5} \\
\bottomrule
\end{tabular}
\end{minipage}
\hfill
\begin{minipage}[t]{0.50\textwidth}
\centering
\caption{\textbf{DML loss-agnostic ablation} on CUB-200-2011 and FGVC-Aircraft (R@1, \%). \textit{bare}: DML loss alone; \textit{SAGA}: same DML loss combined with our GRPO + KL alignment. \textbf{SAGA w/ PF} is the headline configuration of the main paper. Baselines: \textbf{PA} = Proxy Anchor~\citep{proxy_anchor}, \textbf{MS} = Multi-Similarity~\citep{multisimilarity_dml}, \textbf{PF} = Potential Field~\citep{bhatnagar2025potentialfield}.}
\label{tab:ablation_dml}
\small
\setlength{\tabcolsep}{4pt}
\vspace{2pt}
\begin{tabular}{lcccc}
\toprule
& \multicolumn{2}{c}{\textbf{CUB-200} R@1} & \multicolumn{2}{c}{\textbf{Aircraft} R@1} \\
\cmidrule(lr){2-3} \cmidrule(lr){4-5}
\textbf{Method} & bare & SAGA & bare & SAGA \\
\midrule
MS  & 77.8 & 86.1          & 72.5 & 82.3 \\
PA  & 79.5 & 87.3          & 73.1 & 82.7 \\
PF  & 81.6 & \textbf{87.9} & 77.4 & \textbf{83.5} \\
\bottomrule
\end{tabular}
\end{minipage}

\end{table}

\textbf{Loss component analysis:}
Table~\ref{tab:ablation_components} isolates the contribution of each auxiliary loss on top of the PF baseline (PF = Potential Field~\citep{bhatnagar2025potentialfield}). Adding the GRPO term improves R@1 by $\mathbf{5.4}$\% on CUB-200-2011 and $\mathbf{4.9}$\% on FGVC-Aircraft, while the KL alignment term alone yields a much smaller $0.5$\% / $0.7$\% gain on the same datasets, confirming that the GRPO signal contributes the bulk of the attribute-aware supervision and KL plays a complementary, narrower role of supervising the pooler attention. Combining all three losses (\textbf{SAGA}) gives the strongest configuration, exceeding the PF baseline by $\mathbf{6.3}$\% on CUB-200-2011 and $\mathbf{6.1}$\% on FGVC-Aircraft. 

\textbf{DML loss-agnostic ablation:}
Table~\ref{tab:ablation_dml} replaces the PF term inside SAGA with two alternative DML losses, PA (Proxy Anchor~\citep{proxy_anchor}) and MS (Multi-Similarity~\citep{multisimilarity_dml}). All three SAGA variants substantially exceed their bare-DML baselines: on CUB-200-2011 the SAGA gains over the bare DML loss are $\mathbf{8.3}$\%, $\mathbf{7.8}$\%, and $\mathbf{6.3}$\% for MS, PA, and PF respectively, with comparable or larger gains of $\mathbf{9.8}$\%, $\mathbf{9.6}$\%, and $\mathbf{6.1}$\% on FGVC-Aircraft. The consistency of the gains across DML losses confirms that SAGA is DML loss-agnostic. 

\textbf{Prompt sensitivity:}
To separate the contribution of the attribute vocabulary from generic MLLM-oracle access, we re-run \texttt{+GRPO} with the per-dataset attribute list removed (generic variant in Appendix~\ref{app:prompt_generic}); the KL term is omitted because its distillation target (attention pooled over named-attribute spans) is undefined without an attribute vocabulary. The italicised \textit{(generic prompt)} row in Table~\ref{tab:ablation_components} shows the GRPO lift over PF drops from $+5.4 / +4.9$ to $+2.5 / +2.2$ R@1 on CUB-200-2011 / FGVC-Aircraft, a recovery of roughly $45\%$. A generic MLLM oracle therefore accounts for about half of the GRPO gain, and the attribute vocabulary contributes the remaining, confirming that attribute-aware reasoning is a meaningful and quantifiable component beyond a generic LLM-as-oracle baseline.

Additional ablations over embedding dimension and MLLM used are reported in Appendix~\ref{app:additional_ablations}.

\subsection{Attention Analysis}
\label{sec:attention_analysis}
\label{sec:qualitative} 

\begin{figure}[t]
    \centering
    \begin{tabular}{@{}c@{}}
        \includegraphics[width=\textwidth]{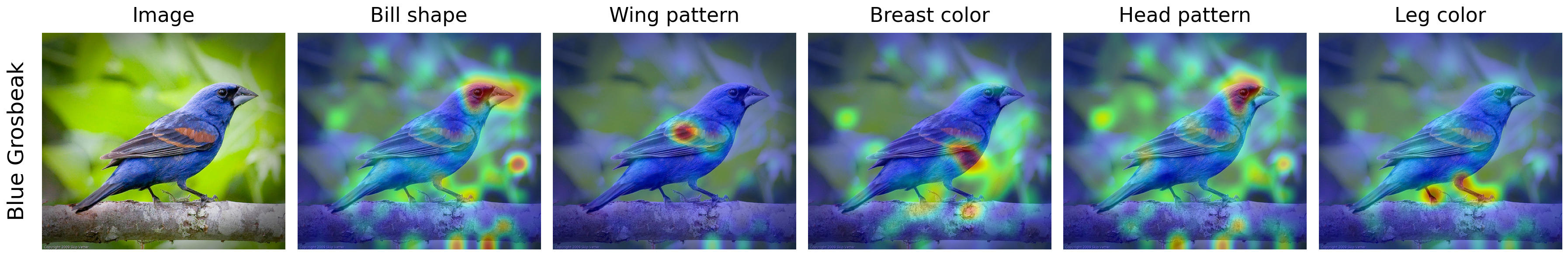} \\[2pt]
        \includegraphics[width=\textwidth]{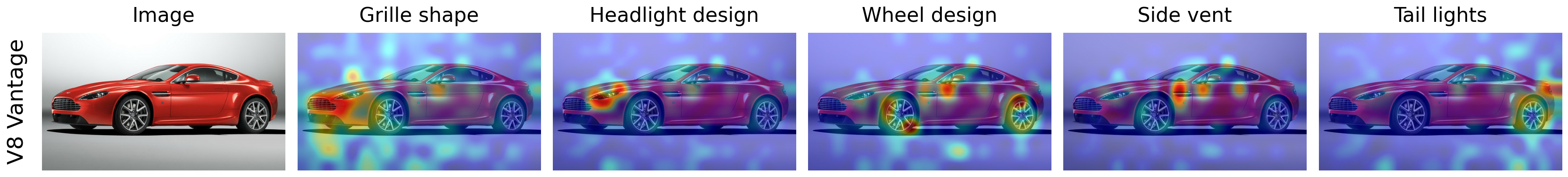}
    \end{tabular}
    \caption{\textbf{MLLM supervisor attention over named attributes (KL target).} For a held-out CUB-200-2011 query (top, Blue Grosbeak) and a Cars-196 query (bottom, V8 Vantage), we overlay the MLLM's attention pooled over the reasoning tokens that name each attribute. For the bird, attention localizes on the bill, wing, breast, head, and legs as each attribute is named; for the car, on the grille, headlights, wheels, side vent, and tail lights. These per-attribute spatial maps are exactly the targets that the KL alignment term distills into the vision pooler.}
    \label{fig:attention_overlays}
\end{figure}

Figure~\ref{fig:attention_overlays} visualizes the MLLM supervisor attention that the KL alignment term distills, on two held-out queries: a CUB-200-2011 Blue Grosbeak and a Cars-196 V8 Vantage. For each attribute the MLLM names in its discriminative-reasoning trace, we pool the supervisor's attention over the tokens corresponding to that attribute and overlay the result on the input image. In every column the mass concentrates on the named attribute region rather than on the bird or car as a whole. This is direct visual evidence that the supervisor signal is attribute-resolved, not a coarse object-vs-background prior, and motivates the per-attribute KL loss in Sec.~\ref{sec:method}: by aligning the pooler's attention with these maps, the vision encoder inherits the attribute-level spatial discrimination that drives the retrieval gains in Table~\ref{tab:main_results}. Retrieval-level qualitative comparisons (top-$k$ images per query, with correct/incorrect class borders, including failure cases) are deferred to Appendix~\ref{app:qualitative_gallery}.

\section{Limitations}
\label{sec:limitations}
Training under our framework is slower than standard DML, as each contributing pair requires $G$ rollouts through the frozen MLLM and a differentiable replay through the language backbone. The added cost is paid only at training time; inference uses the vision encoder and pooler alone and is identical in cost to a vanilla DML pipeline. The framework also presumes a supervisor capable of following the structured comparison prompt and resolving a non-trivial fraction of pairs correctly, since GRPO produces gradient signal only when rollouts disagree on the verdict. Open-weight MLLMs that we use meet this requirement on standard fine-grained benchmarks.

\section{Conclusion}
\label{sec:conclusion}

We introduced SAGA, a framework that turns a frozen MLLM into a training-time supervisor for the vision encoder of a retrieval system. Where class-label DML reduces a pair to a scalar that acts on every embedding direction in unison, GRPO over the MLLM's verdict yields a gradient whose group-normalized advantages cancel on tokens the rollouts agree on and concentrate on the discriminating ones, routing signal into precisely the directions that resolve the attributes the supervisor used to judge. A KL term distills the supervisor's attention over its discriminative-reasoning tokens into the pooler, and a standard metric loss shapes the geometry. The MLLM is frozen throughout and discarded at inference, so deployment cost matches a vanilla DML pipeline; on CUB-200-2011, Cars-196, FGVC-Aircraft, and iNaturalist Aves, this lifts Recall@1 by 3 to 6 points over the strongest baselines on the same backbone. We view the binary verdict as the simplest instance of a broader principle, that coarse rewards adjudicated by a reasoning supervisor can carry far more structure into the gradient than they appear to, and see this as a promising lever for representation learning whenever fine-grained annotation is unavailable.


\bibliographystyle{plainnat}
\bibliography{egbib, main}

\newpage
\appendix

\begin{center}
    {\Large \textbf{Supplementary Material for}}\\[6pt]
    {\Large \textbf{Beyond Scalar Distances: Semantic Attribute Gradients from Frozen MLLMs for Visual Embeddings}}
\end{center}

\vspace{6pt}

In this supplementary material, we provide additional information that did not fit in the main paper. We do so in five sections: Sec.~\ref{app:datasets} gives full statistics and licenses for the four image-retrieval benchmarks; Sec.~\ref{app:impl_details} reports implementation and optimization details (training algorithm, optimizer, loss weights, batching, hardware); Sec.~\ref{app:additional_ablations} reports additional ablations omitted from Sec.~\ref{sec:ablations} for space, including ablations over embedding dimension and vision backbone; Sec.~\ref{app:qualitative_gallery} shows top-5 nearest-neighbor rankings on held-out queries with class-correctness color coding; finally, Sec.~\ref{app:prompt} reproduces the structured pair-comparison prompt $T_\text{inst}$ template used by the supervisor MLLM, with per-dataset substitutions and the attribute vocabularies that instantiate the template for the four datasets in our main experiments.

\section{Dataset Details}
\label{app:datasets}

We provide additional details for each of the four fine-grained image retrieval benchmarks used in our main experiments. All datasets are publicly available under their original licenses; we use them solely for non-commercial academic research. Across all benchmarks we follow the standard zero-shot retrieval protocol of~\citet{sopdataset}: classes are partitioned into disjoint training and evaluation halves, and models are evaluated on classes never seen during training. The structured comparison prompt $T_\text{inst}$ used by the supervisor MLLM and the per-dataset attribute vocabularies that instantiate it for each dataset above are reported at the end of this supplement in Appendix~\ref{app:prompt}.

\paragraph{CUB-200-2011~\citep{cubdataset}.}
The Caltech-UCSD Birds 200-2011 dataset contains $11{,}788$ images of $200$ bird species ($\approx 59$ images per class). Each image is annotated with $312$ binary attributes spanning $28$ attribute groups (bill shape, plumage color, wing pattern, etc.), $15$ part-location keypoints, and a single bounding box. We use the first $100$ species for training and the remaining $100$ for evaluation. Birds are distinguished by subtle attribute combinations such as bill shape, plumage patterning, wing-bar presence, and eye-ring color, making CUB the canonical benchmark for attribute-aware retrieval.

\paragraph{Cars-196~\citep{carsdataset}.}
The Stanford Cars dataset contains $16{,}185$ images of $196$ car classes defined at the make-model-year level (e.g., \textit{2012 Tesla Model S Sedan}). Following the zero-shot DML convention of~\citet{sopdataset}, the first $98$ classes are used for training and the remaining $98$ for evaluation. Classes are distinguished by external visual cues such as body style, grille and headlight design, side profile, badge placement, and apparent era.

\paragraph{FGVC-Aircraft~\citep{maji2013aircraft}.}
The Fine-Grained Visual Classification of Aircraft dataset~\citep{maji2013aircraft} ships approximately $10{,}000$ images organized hierarchically (manufacturer, family, variant). We retrieve at the variant level using the standard $100$-variant release. Following the same disjoint-class convention as CUB and Cars, we sort variants alphabetically and split the $100$ classes into the first $50$ for training and the remaining $50$ for evaluation, pooling FGVC's own \texttt{trainval} and \texttt{test} image partitions before the class-level split (since our train/eval classes are already disjoint, the original image-level split is irrelevant). Discriminative cues include wing configuration (low / mid / high / T-tail), engine count and mounting position, fuselage profile, and vertical-stabilizer geometry.

\paragraph{iNaturalist 2021 Aves~\citep{vanhorn2021inat}.}
We use the Aves (birds) supercategory from the \texttt{train\_mini} split of the iNaturalist 2021 competition, comprising $1{,}486$ species at $50$ images per species ($\approx 74{,}300$ images total). Following the same disjoint-class protocol, we sort species directories lexicographically (zero-padded iNat category-id prefix) and use the first $743$ species for training and the remaining $743$ for evaluation. Compared to CUB, iNat-Aves covers a substantially broader taxonomic range and contains images captured by the iNaturalist citizen-science community under highly varied conditions (lighting, pose, partial occlusion, cluttered natural backgrounds), making it an open-set fine-grained benchmark much closer to real-world species identification.

\paragraph{Image preprocessing.}
All images are resized to $224 \times 224$ before being passed to the Qwen3-VL-8B vision encoder, matching the pre-training resolution of the base model. We do not crop using bounding-box annotations, so the encoder sees the full image including background context.

\section{Additional Implementation Details}
\label{app:impl_details}

\begin{algorithm}[H]
\caption{SAGA: One Training Step}
\label{alg:training}
\begin{algorithmic}[1]
\REQUIRE Batch stream $\mathcal{B}$, target contributing pairs $K$, group size $G$, loss weights $\lambda_\text{dml}, \lambda_\text{lm}, \lambda_\text{kl}$

\STATE \textbf{Phase A: Embedding \& DML} (per micro-batch)
\STATE $\{z_i\}_{i=1}^B \leftarrow c_\phi(f_\theta(I_i))$; backward $\lambda_\text{dml}\cdot\mathcal{L}_\text{DML}(\{z_i\},\{y_i\})$

\STATE \textbf{Phase B: GRPO Rollouts (no grad, DAPO Dynamic Sampling)}
\STATE Buffer $\mathcal{C}\leftarrow\emptyset$
\WHILE{$|\mathcal{C}| < K$}
    \STATE Sample image pair $(I_a, I_b)$ from $\mathcal{B}$ (refill micro-batches as needed)
    \STATE Sample $G$ completions $\{\hat{Y}^{(g)}\}_{g=1}^G$ from $p_\psi$; parse rewards $\{r^{(g)}\}$, advantages $\{A^{(g)}\}$
    \IF{$\sigma_r > 0$}
        \STATE $\mathcal{C}\leftarrow\mathcal{C}\cup\{(\{\hat{Y}^{(g)}\},\{A^{(g)}\},\{r^{(g)}\})\}$
    \ENDIF
\ENDWHILE

\STATE \textbf{Phase C: Policy Update (with grad)}
\STATE Recompute log-probs through $f_\theta\to p_\psi$ for all rollouts in $\mathcal{C}$ ($c_\phi$ bypassed)
\STATE Compute $\mathcal{L}_\text{GRPO}$ via Eq.~\eqref{eq:grpo_loss} over $\mathcal{C}$ (token-level normalization across the buffer); backward $\lambda_\text{lm}\cdot\mathcal{L}_\text{GRPO}$
\STATE For each rollout in $\mathcal{C}$ with $r^{(g)}=1$: compute $\mathcal{L}_\text{KL}$ via Eq.~\eqref{eq:kl_loss} ($\ell=26$ teacher attention); backward $\lambda_\text{kl}\cdot\mathcal{L}_\text{KL}$

\STATE Gradient clip; optimizer step
\end{algorithmic}
\end{algorithm}

\begin{figure}[H]
\centering
\includegraphics[width=1\linewidth]{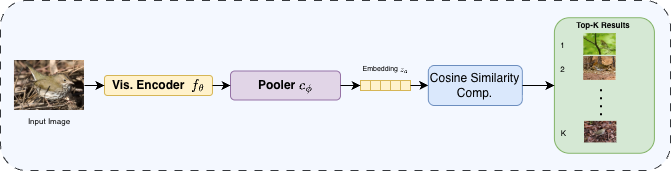}
\caption{Inference-time pipeline. Only $f_\theta$ and $c_\phi$ remain; the frozen MLLM, GRPO rollouts, and KL alignment are dropped. Each image yields a single $\ell_2$normalized embedding used directly for retrieval.}
\label{fig:inference_pipeline}
\end{figure}

\paragraph{Inference.}
At test time the GRPO rollouts, KL alignment, and the frozen MLLM supervisor are discarded; only the vision encoder $f_\theta$ and attention pooler $c_\phi$ remain, producing a single $\ell_2$-normalized embedding per image used directly for nearest neighbor retrieval (Fig.~\ref{fig:inference_pipeline}).

\paragraph{Attention pooler architecture.}
The attention pooler $c_\phi$ is a single-query cross-attention head over the patch tokens $\mathbf{X} \in \mathbb{R}^{N_p \times D}$. A learnable query vector $\mathbf{q} \in \mathbb{R}^{D}$ and a linear key projection $W_k \in \mathbb{R}^{D \times D}$ produce the patch-attention distribution $\boldsymbol{\beta} = \text{softmax}\!\left((\mathbf{q} W_k \mathbf{X}^\top) / \sqrt{D}\right) \in \Delta^{N_p}$, the pooled vector $\boldsymbol{\beta} \mathbf{X}$ is mapped to the embedding dimension $D_e$ by a linear projection, and the resulting embedding is $\ell_2$normalized before being passed to either $\mathcal{L}_\text{DML}$ the retrieval or the distance. The attention weights $\boldsymbol{\beta}$ supervised by $\mathcal{L}_\text{KL}$ (Eq.~\ref{eq:kl_loss}) are exactly the softmax output of this single cross-attention layer; no additional importance-scoring head is used. The DML baselines in Table~\ref{tab:main_results} use mean pooling over patch tokens; we verified on CUB-200-2011 and FGVC-Aircraft that swapping mean for max pooling moves PA and PF R@1 by less than $\sim$0.5\% with no consistent winner, so mean was selected for its conventionality in the DML literature.

\paragraph{Optimisation.}
The vision encoder $f_\theta$ and the attention pooler $c_\phi$ are trained with AdamW at learning rates $2{\times}10^{-5}$ and $1{\times}10^{-4}$ respectively, with cosine annealing over $3$ epochs and a linear warm-up over the first $5\%$ of steps. Gradients are clipped at global norm $1.0$. The frozen language backbone $p_\psi$ receives no gradient updates throughout training.

\paragraph{Loss weights.}
We use $\lambda_\text{dml} = \lambda_\text{lm} = \lambda_\text{kl} = 1$. Each loss term is internally normalized before the weight is applied: $\mathcal{L}_\text{DML}$ is averaged over the batch, $\mathcal{L}_\text{GRPO}$ over generated tokens (Eq.~\ref{eq:grpo_loss}), and $\mathcal{L}_\text{KL}$ over attribute-token positions per image (Eq.~\ref{eq:kl_loss}); these per-loss normalizations bring the gradient magnitudes of the three terms to within roughly an order of magnitude of each other at initialization, so a unit weight on each performs well without further tuning. We did not perform a formal sweep over the weights; preliminary CUB-200 runs at $\lambda_\text{lm}, \lambda_\text{kl} \in \{0.5, 1, 2\}$ produced indistinguishable R@1.

\paragraph{Attention extraction for $\mathcal{L}_\text{KL}$.}
The teacher attention $\boldsymbol{\alpha}$ in Eq.~\ref{eq:kl_loss} is taken from layer $\ell = 26$ of the Qwen3-VL-8B language backbone, head-averaged, and renormalized over the patches of the corresponding image. As discussed in Sec.~\ref{sec:kl_loss}, the last layer is dominated by attention-sink artefacts; layer $\ell = 26$ is the middle-late layer at which sweep visualisations (using the AttWarp~\citep{dalal2025constructive} framework) gave the most spatially grounded attribute-aligned maps.

\paragraph{Batching and GRPO sampling.}
At each micro-batch we draw a class-balanced batch of size $64$ and sample candidate same-/different-class pairs from it; for each pair we roll out $G = 8$ MLLM completions with temperature $0.7$, top-$p$ $0.95$, and at most $1024$ generated tokens. Following DAPO Dynamic Sampling~\citep{yu2026dapo}, we accumulate contributing pairs ($\sigma_r > 0$) across successive micro-batches until $K = 8$ have buffered, then take a single optimizer step over the buffered rollouts. Pairs with $\sigma_r = 0$ contribute neither to $\mathcal{L}_\text{GRPO}$ nor $\mathcal{L}_\text{KL}$ but still receive the DML gradient. $\mathcal{L}_\text{KL}$ is computed only on the rollouts within a buffered pair that received reward $r^{(g)} = 1$. The Qwen3-VL-8B supervisor produces well-formed JSON essentially always, so a fallback parser was unnecessary.

\paragraph{Hardware.}
All experiments use a single NVIDIA H200 (141\,GB) GPU with bfloat16 mixed precision.

\section{Additional Ablations}
\label{app:additional_ablations}

This section reports extended ablations and per-dataset breakdowns for the analyses presented in Sec.~\ref{sec:ablations} of the main paper. Unless otherwise stated, all experiments follow the setting of Sec.~\ref{sec:setup} of the main paper: the Qwen3-VL-8B vision tower with our learned attention pooler, AdamW with the schedule and hyperparameters reported in Appendix~\ref{app:impl_details}, GRPO group size $G = 8$ and DAPO target $K = 8$ contributing pairs per step, and the zero-shot retrieval evaluation protocol with Recall@$K$ and NMI on held-out classes. Each subsection below extends a specific ablation from Sec.~\ref{sec:ablations} of the main paper to additional datasets, hyperparameter ranges, or design choices that did not fit in the main text.

\subsection{Lower-Dimensional Embeddings}\label{app:abl_dim}

\textbf{Context:} The main paper uses embeddings of dimension $d = 4096$. In storage- or compute-constrained retrieval settings (\emph{e.g.}\ on-device species recognition, large-scale gallery indexing) lower-dimensional embeddings are preferable. We verify that SAGA's gain over the baselines is preserved when the embedding is compressed to $d \in \{512, 128\}$, with $d = 512$ matching the standard DML choice in prior work and $d = 128$ being a more aggressive compression target.

\textbf{Experiment:} We re-train SAGA and the PotentialField~\citep{bhatnagar2025potentialfield} baseline, and evaluate the zero-shot encoder, at $d \in \{512, 128\}$, otherwise following the standard setting of Sec.~\ref{sec:setup}. Results are reported on CUB-200-2011 and FGVC-Aircraft.
\begin{table}[h]
\centering
\caption{\textbf{Lower-dimensional embeddings.} R@1 and R@4 (\%) at $d = 512$ and $d = 128$ on CUB-200-2011 and FGVC-Aircraft (the main paper uses $d = 4096$; see Sec.~\ref{sec:setup}). The method ordering is preserved at both compressed dimensions.  \textbf{PF} = Potential Field~\citep{bhatnagar2025potentialfield}.}
\label{tab:ablation_dim}
\small
\setlength{\tabcolsep}{4pt}
\renewcommand{\arraystretch}{1.1}
\resizebox{\textwidth}{!}{%
\begin{tabular}{l cccc cccc}
\toprule
& \multicolumn{4}{c}{$d = 512$} & \multicolumn{4}{c}{$d = 128$} \\
\cmidrule(lr){2-5}\cmidrule(lr){6-9}
& \multicolumn{2}{c}{CUB-200-2011} & \multicolumn{2}{c}{FGVC-Aircraft} & \multicolumn{2}{c}{CUB-200-2011} & \multicolumn{2}{c}{FGVC-Aircraft} \\
\cmidrule(lr){2-3}\cmidrule(lr){4-5}\cmidrule(lr){6-7}\cmidrule(lr){8-9}
\textbf{Method} & R@1 & R@4 & R@1 & R@4 & R@1 & R@4 & R@1 & R@4 \\
\midrule
Pre-trained backbone & $75.0$ & $91.6$ & $50.5$ & $74.1$ & $71.7$ & $90.2$ & $45.8$ & $70.2$ \\
PF                   & $79.4$ & $91.8$ & $76.3$ & $92.2$ & $78.3$ & $91.5$ & $75.5$ & $91.5$ \\
\textbf{SAGA}        & \textbf{86.5} & \textbf{93.1} & \textbf{80.1} & \textbf{93.3} & \textbf{83.8} & \textbf{92.7} & \textbf{79.5} & \textbf{91.7} \\
\bottomrule
\end{tabular}
}
\end{table}

\textbf{Results:} Table~\ref{tab:ablation_dim} reports R@1 and R@4 at $d \in \{512, 128\}$ on CUB-200-2011 and FGVC-Aircraft. The method ordering mirrors Table~\ref{tab:main_results} (which uses the main paper's $d = 4096$): SAGA retains its R@1 margin over PF at both compressed dimensions, indicating that the attribute-aware GRPO signal continues to deliver gains in the small-embedding regime relevant to deployment.


\subsection{Vision Backbone Transfer}\label{app:abl_backbone}

\textbf{Context:} The main paper uses the Qwen3-VL-8B~\citep{bai2025qwen3vl} vision tower throughout. To test whether SAGA's gain transfers beyond a single MLLM family, we re-train the pipeline with the vision tower swapped to InternVL3.5-8B~\citep{chen2024internvl}, keeping the attention pooler, GRPO, and KL alignment unchanged.

\textbf{Experiment:} We train SAGA and the PotentialField~\citep{bhatnagar2025potentialfield} baseline on CUB-200-2011 using the InternVL3.5-8B vision tower as the encoder $f_\theta$. Note that this means that supervisor LM $p_\psi$ is also the InternVL3.5-8B language backbone in these runs. We additionally report the zero-shot retrieval recall of the InternVL3.5-8B encoder (no fine-tuning) as a no-training reference. Due to compute constraints, this transfer study is limited to CUB-200-2011.
\begin{table}[h]
\centering
\caption{\textbf{Vision backbone transfer.} R@1, R@2, R@4, and R@8 (\%) on CUB-200-2011 when the vision tower of $f_\theta$ is swapped from Qwen3-VL-8B (main paper) to InternVL3.5-8B~\citep{chen2024internvl}. For the trained runs, the supervisor LM $p_\psi$ is the InternVL3.5-8B language backbone. \textbf{PF} = Potential Field~\citep{bhatnagar2025potentialfield}.}
\label{tab:ablation_backbone}
\small
\setlength{\tabcolsep}{6pt}
\renewcommand{\arraystretch}{1.1}
\begin{tabular}{lcccc}
\toprule
\textbf{Method} (encoder = InternVL3.5-8B) & R@1 & R@2 & R@4 & R@8 \\
\midrule
Zero-shot       & 50.8 & 64.8 & 76.4 & 85.8 \\
PF only         & 78.0 & 86.2 & 91.8 & 95.1 \\
\textbf{SAGA}   & \textbf{80.1} & \textbf{87.6} & \textbf{92.3} & \textbf{95.6} \\
\bottomrule
\end{tabular}
\end{table}

\textbf{Results:} Table~\ref{tab:ablation_backbone} reports R@1 through R@8 on CUB-200-2011 for the zero-shot encoder, the PF baseline, and SAGA under the alternate encoder/MLLM combination. The SAGA gain over PF persists with the swapped backbone, indicating that the attribute-aware GRPO signal is not tied to a single MLLM family.

\begin{figure}[t]
    \centering
    \includegraphics[width=0.95\textwidth]{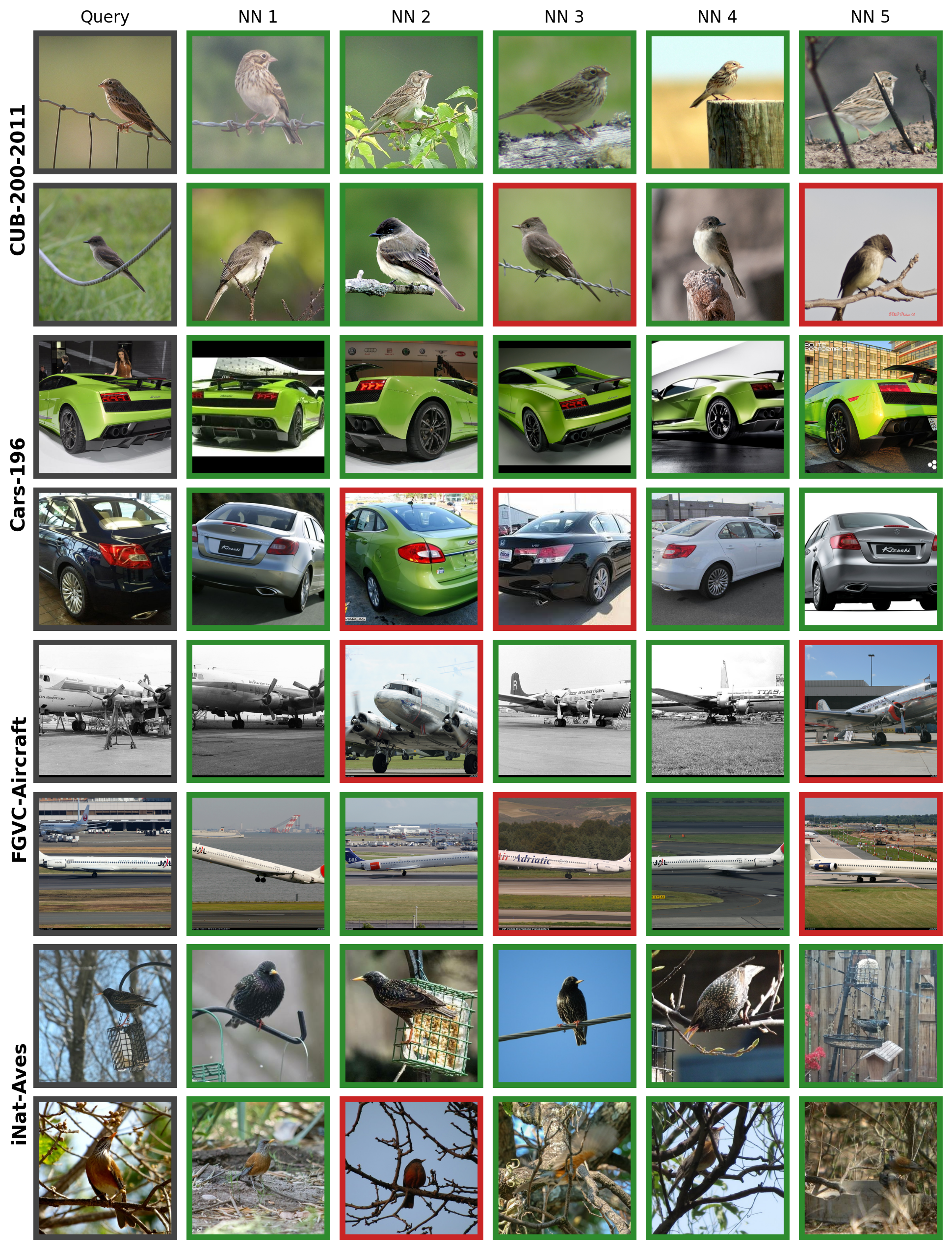}
    \caption{\textbf{Qualitative retrieval gallery on held-out test classes.} Two queries per dataset (rows), top-5 nearest neighbors under the SAGA embedding in descending cosine similarity (columns 2 to 6). The leftmost column is the query (neutral border). \textcolor{green}{Green} borders mark same-class retrievals; \textcolor{red}{red} borders mark cross-class errors.}
    \label{fig:qualitative_gallery}
\end{figure}

\section{Qualitative Retrieval Gallery}\label{app:qualitative_gallery}

\textbf{Context:} We complement the attention-overlay analysis of Sec.~\ref{sec:attention_analysis} with retrieval-level qualitative results: top-5 nearest-neighbor rankings produced by the SAGA embedding on held-out test images.

\textbf{Experiment:} For each of the four benchmarks we report two held-out query images, both drawn from the standard zero-shot retrieval split (classes disjoint from training). To avoid both trivial wins (queries surrounded by easy same-class neighbors) and degenerate cases (queries whose image content is dominated by background), we stratify the candidate pool by the SAGA top-5 hit count: the first row per dataset is drawn from queries whose SAGA top-5 contains four or five same-class neighbors (\emph{clean}), and the second from queries whose top-5 contains one to three same-class neighbors (\emph{informative}). For each query we display the original image (leftmost column, neutral border) followed by its five nearest neighbors in descending cosine similarity, with \textcolor{green}{green} borders for same-class retrievals and \textcolor{red}{red} borders for cross-class errors. All embeddings are $\ell_2$-normalized, and the query is excluded from its own retrieval set.

\textbf{Results:} On the \emph{clean} rows (Fig.~\ref{fig:qualitative_gallery}) SAGA returns same-class neighbors that are also visually consistent with the query. On the \emph{informative} rows the wrong-class neighbors are visually plausible (similar pose, color, or silhouette), so the residual errors sit at the boundary between visually adjacent classes rather than across coarse categories.

\section{Prompts and Attributes}
\label{app:prompt}

\subsection{Prompt Template}

The supervisor MLLM is queried with a structured pair-comparison prompt $T_\text{inst}$ that asks it to (i) describe each of the two input images along a fixed list of visual attribute groups, (ii) summarize the key visual differences between the two images, and (iii) emit a same/different verdict in JSON. The verdict field of the JSON is parsed by the GRPO reward function (Sec.~\ref{sec:method}) to produce the binary reward $r \in \{0, 1\}$ used in the group-relative advantage estimate.

We use the same prompt structure across all four datasets, parameterized by (i) the expert role assumed by the model, (ii) the item word for the photographed object, (iii) the verdict question, and (iv) the dataset-specific attribute list reported in Sec.~\ref{app:prompts_all} below. The generic template is reproduced verbatim below, exactly as it appears in our codebase; per-dataset substitutions are given in Table~\ref{tab:prompt_params}.

\begin{quote}
\small
\begin{verbatim}
You are assisting an {EXPERT_NOUN} in identifying {WHAT_TO_IDENTIFY}
from photographs.

{EXPERT_PREFIX} specialists use the following visual attributes to
distinguish between {TARGET_PLURAL}:
{attr_list}

You are given two {item} photographs ({ITEM} 1 and {ITEM} 2).

Please do the following:

1. **Describe {ITEM} 1**: For each of the attributes listed above,
   describe what you observe in {ITEM} 1. Use natural, concise
   language (e.g. "{EXAMPLE_DESCRIPTION}").

2. **Describe {ITEM} 2**: Do the same for {ITEM} 2.

3. **Key Differences**: Summarize the most important visual
   differences between the two {item_plural}. Focus on the
   attributes that would help a specialist tell them apart.

4. **{VERDICT_LABEL} Prediction**: Based on your observations, are
   these two {item_plural} the {VERDICT_QUESTION}?

Respond in JSON with the following structure:
{
  "{item}_1": {
    "<attr_1>": "...",
    "<attr_2>": "...",
    ... (one entry per attribute)
  },
  "{item}_2": { ... same attribute keys ... },
  "differences": "key visual differences between the two {item_plural}",
  "confidence": "high", "medium", or "low",
  "reasoning": "one-sentence justification based on the attributes",
  "{VERDICT_KEY}": "yes" or "no"
}
\end{verbatim}
\end{quote}

\begin{table}[h]
\centering
\caption{\textbf{Per-dataset parameterisation of the comparison prompt template.} Substituting each column's values into the placeholders of the template above yields the exact prompt used for that dataset. CUB-200-2011 and iNaturalist Aves share a single prompt (both are bird benchmarks). The verdict key is the JSON Boolean field whose value is parsed into the binary GRPO reward.}
\label{tab:prompt_params}
\small
\setlength{\tabcolsep}{4pt}
\renewcommand{\arraystretch}{1.15}
\resizebox{\textwidth}{!}{%
\begin{tabular}{llll}
\toprule
\textbf{Placeholder} & \textbf{CUB-200-2011 / iNat-Aves} & \textbf{Cars-196} & \textbf{FGVC-Aircraft} \\
\midrule
\texttt{\{EXPERT\_NOUN\}}         & ornithologist          & automotive expert         & aviation expert \\
\texttt{\{WHAT\_TO\_IDENTIFY\}}   & bird species           & vehicle make, model, and year & aircraft make, model, and variant \\
\texttt{\{EXPERT\_PREFIX\}}       & Bird                   & Automotive                & Aviation \\
\texttt{\{TARGET\_PLURAL\}}       & species                & vehicle models            & aircraft variants \\
\texttt{\{ITEM\} / \{item\}}      & Bird / bird            & Car / car                 & Aircraft / aircraft \\
\texttt{\{item\_plural\}}         & birds                  & cars                      & aircraft \\
\texttt{\{EXAMPLE\_DESCRIPTION\}} & short, cone-shaped, dark grey & four-door sedan, mid-size, chrome horizontal-slat grille & twin turbofan, under-wing, swept low wing, t-tail \\
\texttt{\{VERDICT\_LABEL\}}       & Same Species           & Same Model                & Same Variant \\
\texttt{\{VERDICT\_QUESTION\}}    & same species           & same make and model       & same make, model, and variant \\
\texttt{\{VERDICT\_KEY\}}         & \texttt{same\_species} & \texttt{same\_model}      & \texttt{same\_variant} \\
\midrule
\textbf{\# attributes}            & 28                     & 17                        & 17 \\
\bottomrule
\end{tabular}%
}
\end{table}

\subsection{Per-Dataset Attribute Vocabularies}
\label{app:prompts_all}

The supervisor MLLM is asked to describe each input image along a fixed list of visual attribute groups before producing its same/different verdict. The list of attribute groups is dataset-specific and was chosen to capture the visual cues that domain specialists actually use to discriminate at the relevant taxonomic level: species for the two bird benchmarks, make and model for Cars-196, and variant (\emph{e.g.}\ Boeing 737-700 vs.\ 737-800) for FGVC-Aircraft. The full per-dataset vocabularies are listed below; together with the prompt template above they fully specify the input passed to the supervisor.

\paragraph{CUB-200-2011~\citep{cubdataset} and iNaturalist Aves~\citep{vanhorn2021inat}.}
For both bird benchmarks we use the same $28$-attribute vocabulary, derived by collapsing CUB's $312$ binary attributes into $28$ groups: \textit{bill shape, bill length, bill color, head pattern, crown color, forehead color, eye color, nape color, throat color, breast color, breast pattern, belly color, belly pattern, back color, back pattern, upperparts color, wing color, wing pattern, wing shape, tail shape, tail pattern, upper tail color, under tail color, underparts color, shape, size, primary color, leg color}.

\paragraph{Cars-196~\citep{carsdataset}.}
$17$ attribute groups, chosen to discriminate at the make-and-model level: \textit{body style, number of doors, front grille, headlights, front bumper, side profile, roofline, greenhouse, wheels, fenders and wheel arches, rear lights, rear bumper, exhaust, badging, overall proportions, apparent era, paint finish}.

\paragraph{FGVC-Aircraft~\citep{maji2013aircraft}.}
$17$ attribute groups, chosen to disambiguate variants (e.g.\ Boeing 737-700 vs.\ 737-800), not just manufacturers or families: \textit{number of engines, engine type (turbofan / turboprop / piston / jet), engine mount position (under-wing / rear-fuselage / tail / in-wing), engine nacelle shape and size, wing configuration (high / mid / low mounted), wing planform (swept / straight / delta / variable), winglets or wingtip shape, tail configuration (conventional / t-tail / cruciform / v-tail), vertical stabilizer shape, fuselage length and proportions, nose shape, cockpit window layout, cabin window count and spacing, landing gear layout, overall size class (light / regional / narrow-body / wide-body), apparent era, livery and markings}.

\subsection{Generic Comparison Prompt}
\label{app:prompt_generic}

The prompt-sensitivity ablation in Sec.~\ref{sec:ablations} replaces the structured prompt above with the generic variant below: the expert role and the attribute list are dropped, and the model is asked only for a free-form difference description and the same/different verdict.
The verdict key is preserved, so the GRPO reward parser operates without modification. We use the same placeholder convention as the structured template above; per-dataset substitutions are read from Table~\ref{tab:prompt_params}. 

\begin{quote}
\small
\begin{verbatim}
You are given two {item} photographs ({ITEM} 1 and {ITEM} 2).

Describe the most important visual differences between the
two {item_plural}, then decide whether they are the
{VERDICT_QUESTION}.

Respond in JSON with the following structure:
{
  "differences": "key visual differences between the two {item_plural}",
  "reasoning": "one-sentence justification",
  "{VERDICT_KEY}": "yes" or "no"
}
\end{verbatim}
\end{quote}


\end{document}